\pdfoutput=1
\documentclass[11pt]{article}

\usepackage[preprint]{acl}

\usepackage{times}
\usepackage{latexsym}

\usepackage[T1]{fontenc}
\usepackage[utf8]{inputenc}

\usepackage{microtype}

\usepackage{inconsolata}

\usepackage{enumerate}
\usepackage{amsthm}
\usepackage{graphicx}
\usepackage{epstopdf}
\usepackage{caption}
\usepackage{color}
\usepackage{enumitem}
\usepackage{array}
\usepackage{tabularx}
\usepackage{multirow}
\usepackage{bm}
\usepackage{booktabs}

\usepackage{tcolorbox}
\usepackage{caption}
\usepackage{subcaption}
\usepackage{listings}
\usepackage{wrapfig}
\usepackage{bbding}
\tcbuselibrary{skins}
\usepackage{colortbl}
\usepackage{fontawesome}

\usepackage{authblk}
\tcbuselibrary{breakable}

\usepackage{etoolbox}
\preto{\abstractkeywords}{\nolinenumbers} 

\title{An Effective Framework to Help LLMs Handle Numeric-involved Long-context Tasks}

\author[a]{Yijiong Yu}
\affil[a]{Tsinghua University}

\begin{document}
\maketitle
\begin{abstract}
Large Language Models (LLMs) have demonstrated remarkable capabilities in handling long texts and have almost perfect performance in traditional retrieval tasks. However, their performance significantly degrades when it comes to numerical calculations in the long-context. Numeric-involved long-context tasks typically cannot be addressed by current LLMs in normal settings due to their inherent limitations in simultaneously handling complex and massive information. Some CoT like prompting methods can improve accuracy but demands massive output tokens, which is costly and slow. To address this issue, we propose a workflow, which decompose a numeric-involved long-context task into 4 low-level subtasks: judging, extracting and processing with code and conclusion. The former 2 subtasks is relatively simple, which allows us to use smaller models for efficiently processing long context. When numerical calculations are required, we use code generated by LLMs to avoid the disadvantage of LLM not being good at calculations. The results in 2 numeric-involved long-context benchmarks demonstrate our workflow can not only improve accuracy, but also significantly reduce the cost of API calls.

\end{abstract}

\section{Introduction}

In the past year, long-context language models (LCLMs) such as GPT-4o-128k \citep{openai_gpt-4_2023} and Gemini-1.5-1000k \citep{gemini_team_gemini_2024} have surged in popularity, raising questions about their efficacy in handling extended context tasks. While various LCLMs have demonstrated excellent long-context retrieval ability by passing the ``Needle in a Haystack'' test \citep{gkamradt_llmtest_needleinahaystack_2023} in over 100k context length, benchmarks like Loogle \citep{li_loogle_2023} and Loong \citep{wang_leave_2024} have highlighted their shortcomings in more complex tasks.

Looking throughout those long-context evaluation works, it can be clearly observed that, though LLMs' performance varies in different types of long-context tasks and there is gap between small and large models, all the models generally perform very badly on numeric-involved long-context tasks. Numeric-involved means the task requires some arithmetic operations, such as comparison, sorting or counting. An example of a numeric-involved (it involves comparing the ages) long-context task is as follows:

\begin{tcolorbox}[colback=white,]
\textbf{The long context:}\\
Hallie Turner is a 21 years old student, graduated from ......\\
The student named Sonali Jain is graduated from ...... His age is 25 ......\\
Jack likes basketball and ......\\
\tcbline
\textbf{Question:}\\
Which student is the oldest?
\end{tcolorbox}

According to the results of previous researches \citep{wang_leave_2024,li_loogle_2023,yu_hyper-multi-step_2024}, as long as involving some even very low-level numerical calculations, the long-context task become very challenging for most LLMs. Though this phenomenon has been discovered in previous evaluations \citep{wang_leave_2024,li_loogle_2023,yu_hyper-multi-step_2024}, however, there is currently no research targeting at solving this type of problems.

Just like using Chain-of-Thoughts \cite{wei_chain--thought_2023} can greatly help LLMs better solve math problems, using some CoT like prompting methods, such as let LLMs first carefully analyze the data in the context and then answer \citep{yu_hyper-multi-step_2024}, can indeed help improve accuracy. However, due to the potential for massive amounts of data to be analyzed in the long context, such methods will have to make LLMs generate numerous tokens for analyzing, which cause the inference speed extremely slow and generate huge API call costs.

\begin{figure*}[h]
    \centering
    \includegraphics[width=1\linewidth]{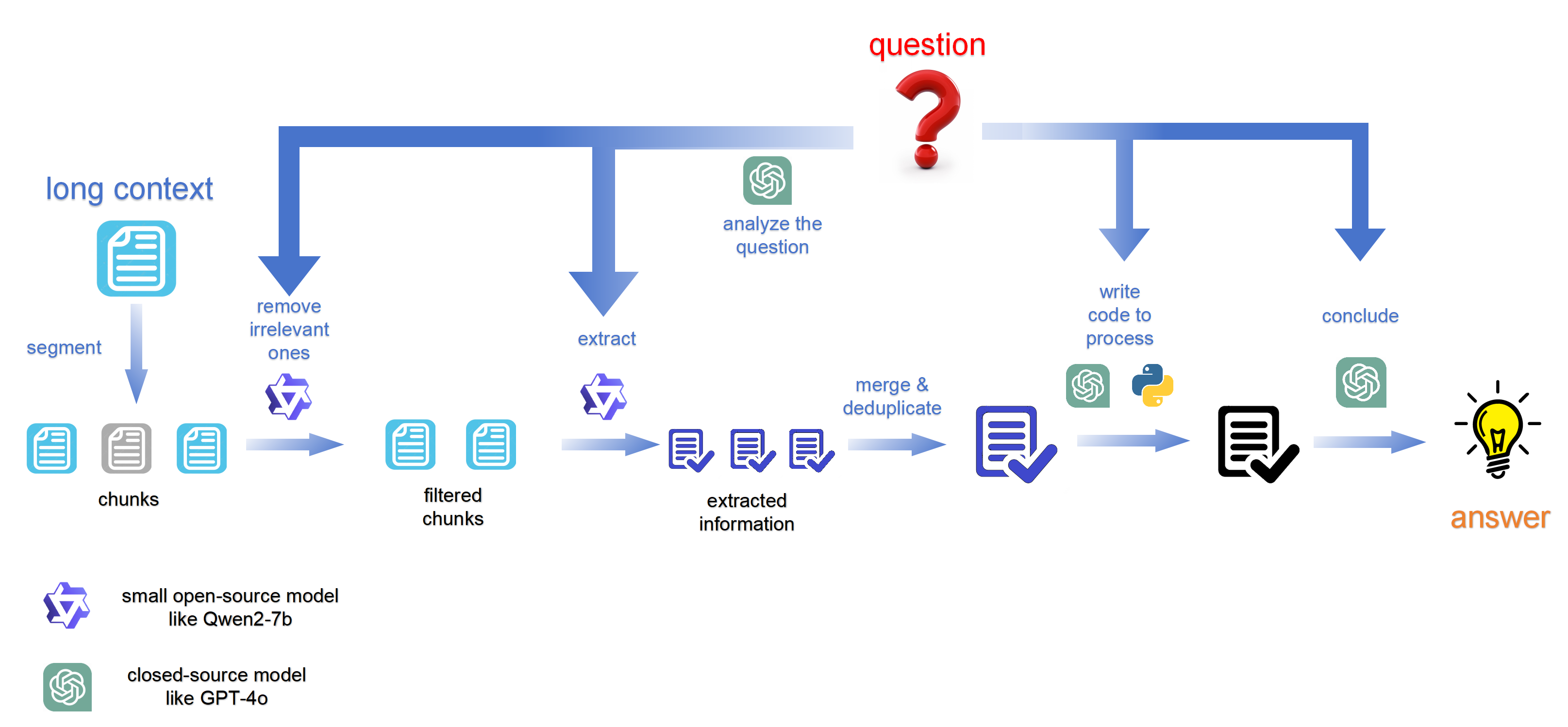}
    \caption{The structure of our method for numeric-involved long-context tasks.}
    \label{fig:structure}
\end{figure*}

Intuitively, a numeric-involved long-context task can be seperated into at least 2 steps: extracting data and analyzing data. Merely extracting data which has been presented in the context is a low-level long-context task, which is not hard for most long-context language models, even small ones like Qwen2.5-1.5b-instruct \cite{yang_qwen2_2024}. But when analyzing data, due to the poor arithmetic ability of LLMs, they often fail, that is why previous researches proposed to prompt LLMs to write code and execute it to solve arithmetic problems.

Therefore, considering the above two characteristics, we propose a more efficient and economic methods, which decomposes a numeric-involved task into multiple low-level steps, and uses small models to offload the pressure of extracting information from long texts, which can avoid the cost of using closed-source models. And when involving calculating the extracted data, we prompt LLMs to write code and execute it with a external code interpreter to process the data. Finally, the LLM use the processed data to conclude a final answer. Figure \ref{fig:structure} shows the overall process of our method.

We test our method on 2 numeric-involved long-context benchmarks, Loong \cite{wang_leave_2024} and difficult-retrieval \cite{yu_hyper-multi-step_2024}. The results demonstrate our method can not only improve accuracy, but also greatly save money. It is an effective, efficient and economic framework, and can be easily applied to long-context QA tasks.

% \section{related works}
% Previous benchmarks often involved numeric-involved long-context task, but few of them individually researched this type of problem, despite most LLMs still perform very badly in it.

\begin{table*}[t]
    \centering
    \begin{tabular}{c|cc|ccc}
    \toprule
    \multirow{2}{*}{\textbf{Method}} & \multicolumn{2}{c|}{\textbf{Dense}} & \multicolumn{3}{c}{\textbf{Sparse}} \\
    \cmidrule(lr){2-3} \cmidrule(lr){4-6}
    & \textbf{Accuracy(\%)} & \textbf{Cost(\$)} & \textbf{Acc comp.(\%)} & \textbf{Acc clus.(\%)} & \textbf{Cost(\$)} \\
    \midrule
    normal prompt & 6.0 & 0.09 & 31.2 & 54.4 & 0.49 \\
    CoT prompt & 28.0 & 0.15 & 41.2 & 84.4 & 0.52 \\
    \midrule
    Ours & 99.0 & 0.01 & 36.6 & 64.2 & 0.01 \\
    \bottomrule
    \end{tabular}
    \caption{Performance of different methods on numerical-dense and numerical-sparse long-context tasks.}
  	\label{tab: merged}
\end{table*}

\section{Method}
Our workflow can be divided into the following steps:

\paragraph{Analyze the question} First, we prompt the LLM to analyze which types of data is needed for the question. We record the data types given by the LLM as auxiliary information for the following procedures. For example, if the question is ``which company has the highest profit?", the data types should be ``company name" and ``profit". The prompt we use is:

\paragraph{Segment} Due to the excessive length of the original context which may challenge a small model, we simply segment it into short chunks with the length of 1000 tokens.
\paragraph{Remove irrelevant parts} We use a light-weight model, Qwen2.5-1.5b-instruct to determine whether the chunk contains relevant information of the required data types of the question. This is a low-level binary classification task, so using a light-weight model is more efficient. If a chunk is classified as irrelevant, it will be removed. 
\paragraph{Segment again} However, too small chunk size may cause the originally linked information be separated into 2 chunks, for example, the company name and its profit appear in 2 chunks, which makes it hard for LLMs to extract a complete piece of information. Therefore, we merge the remaining chunks and again segment it into chunks of a larger size, such as 8k tokens.
\paragraph{Extract data} Then, for each chunk, we use a medium-size model, Qwen2.5-7b-instruct, to extract the required types of data from it. We prompt the model to organize the extracted data into a markdown table. Then we extract the table from the model's response and convert it into a dataframe. Finally, the dataframe extracted from each chunk is concatenated into one dataframe, and we deduplicate it by the element in the first column.
\paragraph{Process data} We provide the LLM with the question and the first a few rows of the extracted dataframe to let it write a python program, which can process the dataframe with necessary operations, and print the answer with the data in the dataframe. 
\paragraph{Conclusion} According to the output result of code execution as well as the source code, the LLM concludes the results and gives the final answer of the question.\\

In our workflow, the large LLM does not need to face the long context, but only need to analyze the question and write code, which consumes relatively very few tokens. Meanwhile, doing some rudimentary and low-level operations based on the original long context relies on light-weight models, which is more efficient. Moreover, the process of processing each chunk separately can be executed in parallel through distributed computing with multiple GPUs, which let the speed increase exponentially. 

Additionally, in our approach, the choice of model is actually very flexible. For example, besides Qwen, we can also choose cheaper API models like gpt-4o-mini as the auxiliary model.

\section{Experiment}

\subsection{Evaluation Dataset}

% And we also consider using deepseek as the auxiliary model.
Because there are currently no benchmarks exclusively designed for numeric-involved long-context tasks, we separate some subset from previous benchmarks, Loong \cite{wang_leave_2024} and difficult-retrieval \cite{yu_hyper-multi-step_2024}.

The dataset from Loong consists of tasks requiring analyzing the numeric in company financial reports, and we call it financial report analyzing. The number of reports in the context is no more than 5, but the context length is very huge, ranging from 40k to 200k. Each financial report contains massive and somewhat disorganized content, making it challenging to for LLMs to extract the key information. The dataset is divided into 2 problem types, comparison and cluster, and each of them contains 90 samples (we only select the English samples).

The dataset from difficult-retrieval requires analyzing the numeric in hundreds of student resumes, but the context length is relatively short, about 20k when there are 300 resumes in the context. There are totally 100 samples.

Base on their characteristics, we call the former dataset as numerical-sparse and the latter as numerical-dense, according to the number of values to be processed in the context. Evaluation on 2 distinct datasets can prove our method is able to adapt to various tasks with distinct data forms and information distributions.

Due to the very low accuracy of simply prompting LLM to directly answer, we compare our methods with another stronger baseline introduced in difficult-retrieval \cite{yu_hyper-multi-step_2024}, which prompts LLMs to examine the information of each item before giving the final answer. We call this method as a CoT like prompting method.

\subsection{Model Settings}

We use a commonly used long-context language model, gemini-1.5-flash \cite{gemini_team_gemini_2024}, as the main model, i.e., the model accounts for analyzing the question, writing code and giving the final answer. We use 2 smaller models, Qwen2.5-7b-instruct and Qwen2.5-1.5b-instruct \cite{yang_qwen2_2024} as low-level auxiliary models. Qwen2.5-7b-instruct and is used to extract the data, and Qwen2.5-1.5b-instruct is used to determine which part of the context is irrelevant with the question.

For all models, we set the temperature to 0 during inference to ensure stable and accurate results. The max generated tokens is set to 4k. When calculating the accuracy, we use GPT-4o \cite{openai_gpt-4_2023} as the judge to make a ``correct'' or ``incorrect'' judgment on the generated answer based on the reference answer.

When calculating the API cost, the price is \$5 per 1 million input tokens and \$15 per 1 million output tokens.

\subsection{Results}

The results in Table \ref{tab: merged} show the accuracy and average cost per sample in the 2 datasets. The results in the numerical-dense dataset demonstrate our method can greatly improve LLMs' answer accuracy in numerical-dense long-context tasks, and the cost of API calls is the lowest. In contrast, prompting method cannot help much, and will cost much more.

As for numerical-sparse tasks, our methods can still improve the accuracy by near 10\%, though prompting methods achieves higher accuracy. However, because our method avoids using API models to process massive tokens, the cost of our method is much lower than traditional methods.

\section{Conclusion}

In this study, we introduced a novel workflow designed to enhance the performance of large language models in numeric-involved long-context tasks. By decomposing these complex tasks into simpler subtasks, judging, extracting, processing with code, and concluding, we leveraged smaller models for efficient long-context processing and utilized LLMs to generate and execute code for accurate numerical calculations. Our method was evaluated on two distinct benchmarks, demonstrating significant improvements in accuracy and substantial cost reductions compared to traditional prompting methods. This approach not only addresses the inherent limitations of LLMs in handling numeric calculations but also provides a scalable and cost-effective solution for long-context tasks involving numerical data.

\section{Limitations}
Our method is applicable to numeric-involved long-context tasks, which means this method cannot be unconditionally applied to all problems, and must be actively activated by users when facing suitable problems. 

In addition, the accuracy of data extraction may vary depending on the choice of the auxiliary model, for example, using a too small model may cause low accuracy despite improving efficiency. But this influence has not been studied.

\bibliography{custom}
\clearpage
\appendix

\section{The prompt used in our workflow}

\begin{tcolorbox}[colback=white,title=Analyze the question,breakable]
\small{{\sffamily
I have a data sheet and a question according to the data sheet: \\
Q:\{question\}\\
Please analyse this question to determine: What is the header of the analysis data table? And which is the primary key?\\
For example, if the question is "Which company has the highest non-current assets?"
then you should guess the data sheet has a header like "Company Name" or "Non-current Assets", and the primary key is "Company Name".\\
Return your answer in the format of this example:\\
\textasciigrave \textasciigrave \textasciigrave  header:\\
| Company Name | Non-current Assets |\\
\textasciigrave \textasciigrave \textasciigrave\\
\textasciigrave \textasciigrave \textasciigrave primary key \\
Company Name\\
\textasciigrave \textasciigrave \textasciigrave\\
}}
\end{tcolorbox}

\begin{tcolorbox}[colback=white,title=Remove irrelevant parts,breakable]
\small{{\sffamily
A document fragment:
\{doc\_chunk\}\\ 
Please determine if there is any relevant information in the document fragment about:\\
\{headers\_list\}.\\ 

Even if the information is not complete, as long as there is relevant information, the answer should be "Yes".\\ 

Example:\\ 
document: "It has a profit of 200, and the number of workers is 100."\\ 
about: company name or profit\\ 
answer: Yes, there is information about the profit.\\ 

document: "The Google company is the biggest company in the world."\\ 
about: company name or profit\\ 
answer: Yes, there is information about the company name.\\ 

document: "It has a profit of 200, and the number of workers is 100."\\ 
about: company name or market value\\ 
answer: No, there is no information about the company name or the market value.\\ 

Now, give your answer.\\ 
}}
\end{tcolorbox}

\begin{tcolorbox}[colback=white,title=Extract data,breakable]
\small{{\sffamily
A document fragment (some irrelevant content has been omitted):\\ 
\{doc\_chunk\}\\ 

Please extract the necessary information from the document fragment, according to the table header \\ 
\{table\_header\}\\ 

Return your answer in the format of a markdown table with the above header and add the index column with column name "index", for example:\\ 
\textasciigrave \textasciigrave \textasciigrave table\\ 
| index | name | profit |\\ 
|--|--|--|\\ 
| 1 | company A | 200 |\\ 
| 2 | company B | 500 |\\ 
\textasciigrave \textasciigrave \textasciigrave\\ 
If there is no complete data that can be extracted, please return \\ 
\textasciigrave \textasciigrave \textasciigrave table\\ 
no data\\ 
\textasciigrave \textasciigrave \textasciigrave\\ 
Note that there should not be any empty or uncertain element in the table. That is, make sure the extracted data is accurate. For incomplete or unknown data, please discard this row directly. \\ 
For example, if you find a number indicating the profit, but you cannot determine which company it belongs to, you should discard this row.\\ 
Make sure the table header in your answer is strictly consistent with the given header, even the capitalization.\\ 
When extracting numbers, you should ensure that the numbers do not contain commas or other symbols, and the decimal point should be a period.\\ 
Ensure your answer is in correct markdown format.\\ 
}}
\end{tcolorbox}

\begin{tcolorbox}[colback=white,title=Process data by code,breakable]
\small{{\sffamily
I have a pandas Dataframe which has hundreds of rows, and here is the first a few rows of it displayed in markdown format: \\
\{df\_show\} \\

Here is a question: \\
Q:\{question\} \\

Please write a code snippet to process the data to solve the question, but you do not need to execute the code. \\
In the code, you should import necessary libraries and use \\
\textasciigrave \textasciigrave df=pd.read\_json("./data.json",dtype=False)\textasciigrave \textasciigrave  \\
to load the data.  \\

After processing, use \\
\textasciigrave \textasciigrave to\_json("./data\_p.json")\textasciigrave \textasciigrave  \\
to save the result. \\
Finally, you should organize the processed data into a string named "final\_answer" to let users quickly understand the final result. You should print it and save it to "./code\_answer.txt". \\
Note that all the data types are initially string, so you should convert them to the correct data type such as float in your code if necessary. \\
}}
\end{tcolorbox}

\end{document}